%% file: main.tex
\documentclass{ifacconf}

\include{packages}

\include{commands}

\graphicspath{ {./figures/} }

\begin{document}
\begin{frontmatter}
\vspace{-5mm}
\title{Newtonian and Lagrangian Neural Networks: \\ A Comparison Towards Efficient Inverse Dynamics Identification}

\author[First]{Minh Trinh${}^{1,}$\hspace{-0.5mm}}
\author[Second,Third]{A. René Geist${}^{1,}$\hspace{-0.5mm}}
\author[First]{Josefine Monnet}
\author[First]{Stefan Vilceanu}
\author[Second]{Sebastian Trimpe}
\author[First]{Christian Brecher}

\address[First]{Laboratory of Machine Tools and Production Engineering, RWTH Aachen University, Germany (email: \{m.trinh, c.brecher\}@wzl.rwth-aachen.de)}
\address[Second]{Institute for Data Science in Mechanical Engineering, RWTH Aachen University, Germany (email: trimpe@dsme.rwth-aachen.de)}
\address[Third]{Autonomous Learning, University of T\"ubingen, Germany (email: rene.geist@uni-tuebingen.de)}

\thanks{\textbf{Equal contribution.}}

\begin{abstract}
Accurate inverse dynamics models are essential tools for controlling industrial robots. Recent research combines neural network regression with inverse dynamics formulations of the Newton-Euler and the Euler-Lagrange equations of motion, resulting in so-called \emph{Newtonian neural networks} and \emph{Lagrangian neural networks}, respectively. These physics-informed models seek to identify unknowns in the analytical equations from data. Despite their potential, current literature lacks guidance on choosing between Lagrangian and Newtonian networks. In this study, we show that when motor torques are estimated instead of directly measuring joint torques, Lagrangian networks prove less effective compared to Newtonian networks as they do not explicitly model dissipative torques. The performance of these models is compared to neural network regression on data of a MABI MAX 100 industrial robot.
\end{abstract}


\end{frontmatter}

\section{Introduction}
\label{sec:intro}
\begin{figure}[t]
    \centering
    \includegraphics[width=1\columnwidth]{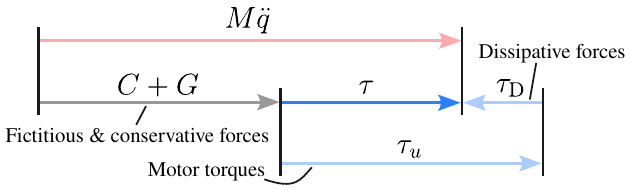}
    \vspace{-5mm}
    \caption{The additive structure of inverse dynamics is visualized as a 1D vector diagram. Inverse dynamics is typically written in terms of the joint torques $\tau$. %
    }
    \label{fig:torque_vectors}
    \vspace{-4mm}
\end{figure}

Inverse dynamics play a crucial role in various robotic applications such as manipulation and machining \citep{nguyen2008learning}.
A robot's inverse dynamics (ID), cf.\,\citep{featherstone2014rigidbody}, describe the robot's joint torques $\tau$ as a function of its generalized coordinates $q\in \Rn$ reading
\begin{equation} \label{eq:RBD_model}
    \tau(q, \dot q, \ddot q) = M(q) \ddot{q} - C(q, \dot{q}) - G(q),
\end{equation}
with the generalized velocity and acceleration vectors $\dot q, \ddot q$, respectively, positive-definite inertia matrix $M(q)$, fictitious forces $C(q, \dot{q})$, and gravitational forces $G(q)$. Typically, $q$ denotes a robot's joint angles. While for readability's sake, we assumed that $\dot q \in \Rn$, the following discussion also applies for the velocity representation having less dimensions than the positions.
The ID can be visualized as a 1D vector diagram as in Figure \ref{fig:torque_vectors} where $\tau$ also denotes the difference between motor torques $\tau_u$ and dissipative torques $\tau_{\mathrm{D}}$.
Eq.\,\eqref{eq:RBD_model} depends on a set of physical parameters $\trbd$ (such as a body's length and mass).
For the sake of brevity, we do not explicitly state the dependency of $M(q)$, $C(q)$, and $G(q)$ on $\trbd$ and the dependency of state variables on the time $t$. 
In practice, \eqref{eq:RBD_model} deviates from torque observations $y$ by a residual
\begin{equation} \label{eq:errors}
    \epsilon(q, \dot q, \ddot q) = y - \tau(q, \dot q, \ddot q).
\end{equation}
This residual in the robot's dynamics model arises from: \emph{(i)} misspecifications of $\trbd$,
\emph{(ii)} inaccurate descriptions of physical phenomena \eg motor dynamics, friction, and elasticities, 
and \emph{(iii)} measurement errors when observing $q$, $\dot q$, $\ddot q$, and $\tau$.

In this study, we aim to identify ID by exploring a recent research branch that aims to reduce the mentioned errors \eqref{eq:errors} by combining neural network regression with two common descriptions of robot dynamics: the Newton-Euler and Euler-Lagrange equations, resulting in \emph{Newtonian networks} and \emph{Lagrangian networks}, respectively. While these approaches are promising, we have not found a comprehensive analysis of their aptitude toward learning inverse dynamics. In this work, we scrutinize these models and then turn to data from an industrial robot to address the question:

\renewenvironment{quote}
  {\begin{trivlist} \setlength\leftskip{0.5cm} \setlength\rightskip{0.5cm}
   \item\relax}
{\end{trivlist}}
  
\begin{quote}
\emph{When learning inverse dynamics with motor torque measurements, how do Newtonian and Lagrangian neural networks compare?}
\end{quote}

We explore this question by comparing Newtonian and Lagrangian networks where joint torques are directly being measured or only motor torque estimates are available.

The contributions of this work are as follows:
\begin{itemize}
    \item we examine the capability of Lagrangian networks to learn errors in joint torques;
    \item we analyze the forces acting in an industrial robot;
    \item we compare Newtonian and Lagrangian neural networks in their proficiency to learn inverse dynamics.
\end{itemize}
As we show in this work, Lagrangian networks can learn friction if joint torques are measured directly. 
However, if motor torques are estimated solely from electric currents, then one must incorporate an additional friction model into Lagrangian networks. In this case, our analysis of the forces occurring in an industrial robot reveals that Newtonian networks compare favorably to Lagrangian networks, if an estimate of the frictionless dynamics is given.

\vspace{1mm}
\subsection{Related work}
Works on robot dynamics identification either identify physical parameters from data, replace physics entirely with data-driven models, or combine both approaches. In what follows, we give a brief overview of recent trends in this field.

\subsubsection{Parameter estimation}
Early works identified the errors \eqref{eq:errors} by estimating physical parameters $\trbd$ from data. Methods that only rely on the estimation of $\trbd$ \eg via linear regression \citep{an1985estimation, siciliano2009modelling} or gradient descent \citep{sutanto2020encoding, lutter2021differentiable} do not mitigate errors due to an inaccurate description of physical phenomena. By neglecting other errors in \eqref{eq:errors}, the estimates of $\trbd$ may acquire physical implausible values \eg negative masses or violate the parallel axis theorem  \citep{traversaro2016identification}. %

\subsubsection{Network regression}
Alternatively, \cite{nguyen2008learning} and \cite{yilmaz2020neural}
deploy neural networks to learn robot dynamics solely from data. %
However, training a neural network to sufficient accuracy requires a large data set which 
has to be collected in real-time while the information contained within the data depends on the collection strategy.

\subsubsection{Physics-informed machine learning}
To improve the models' data efficiency and generalization performance, recent works combine machine learning with physics.
For a detailed discussion on existing works in this field, the reader is referred to \citep{geist2021structured, lutter2023combining}. One can combine a known equation from physics with machine learning in the following manner: \emph{i)} learn the solution to the physics with a neural network while using the physics equation as a regularization term in the network's loss function, and \emph{ii)} combine the physics equation with a network to obtain a model whose predictions are then fed to a loss function. 
The first modeling approach is known under the heading of \emph{physics-informed neural networks} \citep{cuomo2022scientific} and bears the advantage that the neural network's parameter gradients are not directly being transformed by the physics equation. However, using an erroneous physics equation as a regularizer may negatively affect training.
Lagrangian and Newtonian networks follow the second approach and directly combine physics equations with neural networks. In this approach physics errors are approximated directly, while it is often more data efficient to learn errors inside the physics instead of learning transformed errors at the model output. 

Early research in physics-informed machine learning in robotics integrated the linear form of manipulator dynamics \citep{an1985estimation} with Gaussian processes \citep{nguyen2010using, de2011online}. Subsequent advancements introduced models that melded the Newton-Euler equations \citep{geist2020gauss, rath2022physics} and Euler-Lagrange equations \citep{cheng2015learning, giacomuzzo2023embedding} with Gaussian processes. 
Numerous studies shed light on diverse facets of integrating neural networks with the Newton-Euler equations \citep{lutter2021differentiable, djeumou2022neural} and Euler-Lagrange equations \citep{lutter2019lagrangianNN, lutter2019lagrangianNNfuruta, cranmer2020lagrangian}
. Closely, related to the latter are Hamiltonian networks \citep{greydanus2019hamiltonian, finzi2020simplifying} which combine Hamiltonian mechanics with neural networks. Albeit, Hamiltonian mechanics sees infrequent use in robot dynamics identification \citep{ross2016controlling, zada2017application}.

\begin{figure*}[t]
\centering
\includegraphics[width=0.8\textwidth]{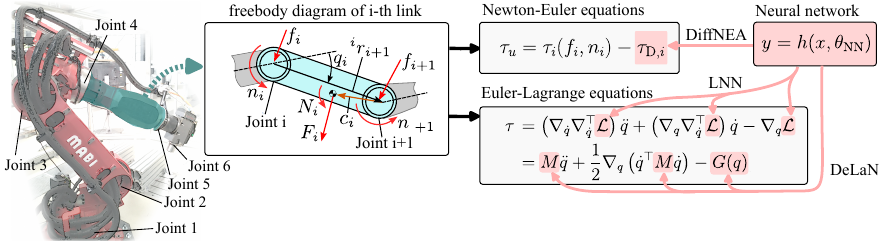}
\vspace{-2mm}
\caption{The inverse dynamics of a MABI MAX 100 shall be learned using physics-informed machine learning. In Newtonian networks (RNEA+MLP), the neural network approximates the residual between joint torque observations and the Newton-Euler equations. In Lagrangian networks starting from the Euler-Lagrange equations, the neural network either replaces the Lagrangian (LNN), or the entries of the inertia matrix inside the Lagrangian (DeLaN).}
\label{fig:overview}
\end{figure*}

\vspace{1mm}
\section{Joint torques are not motor torques}
Our goal is to learn an accurate ID model \eqref{eq:RBD_model}. Here, the term \emph{learning} refers to supervised regression of a differentiable parametric function $y = f(x; \theta)$ via gradient-based optimization using a data set $\mathcal{D}=\{x_i, y_i\}_{i=1}^{N}$ where $N$ denotes the number of data points and $\theta$ the model's parameters. While the inputs to an ID model are typically $x = [q, \dot q, \ddot q]$, we must act with caution when choosing the model's output $y$.
As illustrated in Figure \ref{fig:torque_vectors}, the joint torques are governed by
\begin{equation} \label{eq:torque}
    \tau = \tau_u + \tau_{\text{D}},
\end{equation}
with the motor torques $\tau_u$ and dissipative forces $\tau_{\text{D}}$ where the latter refers to torques due to friction, damping, backlash, and shaft elasticities.

If we directly observe joint torques such that $y:=\tau$ one could approximate it directly as done by \cite{hwangbo2019anymalrobustcontrol}. However, for many robotic systems $\tau$ is not measured and instead one obtains estimates for the motor torques as
\begin{equation} \label{eq:motor_current_est}
    \tau_u = \psi_m K_m I_m
\end{equation}
with the gear ratio $\psi_m$, motor torque constant $K_m$, and measured motor current $I_m$ \citep{siciliano2009modelling}. The case where $y:=\tau_u$ is relevant for feed-forward control where we want to predict $\tau_u$ to be able to track a reference trajectory. The literature on learning ID often derives models that predict $y:=\tau_u$ while the output of the model is being falsely referred to as \emph{``joint torque''}. As we will show, this ambiguity between joint and motor torques can lead to model design decisions that cause subpar performance when resorting to Lagrangian networks.

\section{Newtonian and Lagrangian Networks}
Choosing between Newtonian and Lagrangian networks requires a nuanced understanding of how these models represent robot dynamics. Figure \ref{fig:overview} provides an overview of both approaches with notation drawn from \cite[p. 96]{featherstone2014rigidbody}.
If the same set of generalized coordinates are used then both the Newton-Euler and the Euler-Lagrange equations arrive at identical sets of symbolic equations. However, how these equations are obtained differs significantly. As a consequence, 
if we place a neural network into these equations, then its output undergoes very differing transformations potentially affecting the model's data-efficiency and generalization performance.

\subsection{Lagrangian neural networks}
The Euler-Lagrange equations derive ID \eqref{eq:RBD_model} via the Lagrangian
\begin{equation} \label{eq:Lagrangian}
        \mathcal{L}(q, \dot q) = T(q,\dot q) - V(q) = \frac{1}{2}\dot{q}^{\top} M(q) \dot{q} - V(q),
\end{equation}
with the system's kinetic energy $T(q,\dot q)$, potential energy $V(q)$ and a computational cost of $O(n^2)$. The Euler-Lagrange equations are obtained \eg via the calculus of variations or D'Alembert's principle, reading 
\begin{align}
    \tau &=  \left(\nabla_{\dot{q}}\nabla_{\dot{q}}^{\top} \mathcal{L}\right) \ddot{q} + \left(\nabla_{q}\nabla_{\dot{q}}^{\top} \mathcal{L}\right) \dot{q} - \nabla_{q} \mathcal{L}, \label{eq:euler-lagrange-1} \\
    &= M \ddot{q} + \underbrace{\nabla_{q} \left(\dot{q}^{\transp} M\right) \dot{q} - \frac{1}{2}\nabla_{q} \left(\dot{q}^{\transp} M \dot{q} \right)}_{-C(q,\dot q)} + \underbrace{\nabla_{q} V(q)}_{-G(q)}, \label{eq:euler-lagrange-2}
\end{align}
with $G(q) = - \nabla_{q} V(q)$, and the $i$-th entry in the vector of partial differential operators $\nabla_{\dot{q}}$ being denoted as $(\nabla_{\dot{q}})_i = \frac{\partial}{\partial \dot{q}_i}$.

\subsubsection{DeLaN}
\cite{lutter2019lagrangianNN} combined neural network regression with the Euler-Lagrange equations by replacing $G(q)$ and the entries of the inertia matrix $M$ inside \eqref{eq:euler-lagrange-2} using the neural networks $G_{\text{DLN}}(q; \tnn)$ and $L_{\text{DLN}}(q;\tnn)$. The resulting model is termed \emph{Deep Lagrangian neural network} (DeLaN)
\begin{equation} \label{eq:delan}
    \tcboxmath[left=2pt,right=2pt]{
    \tau = L_{\text{DLN}}^\transp L_{\text{DLN}} \ddot q + \frac{1}{2} \nabla_{q} \left(\dot{q}^{\transp} L_{\text{DLN}}^\transp L_{\text{DLN}} \dot{q} \right) - G_{\text{DLN}}.
    }
\end{equation}
To better model dissipative forces when only observing $\tau_u$, \cite{lutter2019lagrangianNNfuruta} added an analytical friction model to \eqref{eq:delan}. This model achieved almost similar prediction accuracy to linear system identification on data of a Furuta pendulum. Similarly, \cite{gupta2020structured} added a network designed to model dissipative forces to DeLaN which performed slightly better than a multilayer perceptron (MLP).

\subsubsection{LNN}
Alternatively, \cite{cranmer2020lagrangian} replace the Lagrangian in \eqref{eq:euler-lagrange-1} by a neural network $\mathcal{L}_{\text{\lnn}} = h(x; \tnn)$ to arrive at the \emph{Lagrangian neural network} (LNN)
\begin{equation} \label{eq:lnn}
    \tcboxmath[left=2pt,right=2pt]{
    \tau =  \left(\nabla_{\dot{q}}\nabla_{\dot{q}}^{\top} \mathcal{L}_{\text{\lnn}}\right) \ddot{q} + \left(\nabla_{q}\nabla_{\dot{q}}^{\top} \mathcal{L}_{\text{\lnn}}\right) \dot{q} - \nabla_{q} \mathcal{L}_{\text{\lnn}}.
    } 
\end{equation}
\cite{cranmer2020lagrangian} tested LNNs on low-dimensional simulations of conservative dynamical systems where the model outperformed an MLP baseline. We note that during our experiments, it was challenging to find a good weight initialization for LNNs and DeLaN.%

\vspace{1mm}
\subsection{Newtonian neural networks}
A robot's ID \eqref{eq:RBD_model} is typically derived via the recursive Newton-Euler algorithm (RNEA) due to its computational cost of $O(n)$ for kinematic trees. 
The interested reader is pointed to \cite{featherstone2014rigidbody} %
for an in-depth discussion of the RNEA. 

To mitigate parameter errors, \cite{sutanto2020encoding} re-implemented the RNEA in a library for automatic differentiation resulting in the differentiable Newton-Euler algorithm (DiffNEA).  
The DiffNEA also includes a viscous friction-damping model whose parameters were estimated alongside $\trbd$ using data from a KUKA IIWA robot. In this study, the model is significantly more accurate than MLPs and Lagrangian networks.

The DiffNEA is a good model if errors arise solely from a misspecification of $\trbd$. To mitigate errors in the friction torque $\tau_{\mathrm{D}}$, \cite{lutter2020differentiable} replaced the analytical friction model in the DiffNEA with a neural network $h(q,\dot q;\tnn)$, yielding a Newtonian network (\emph{RNEA + MLP})
\begin{equation} \label{eq:newtonian-networks}
    \tcboxmath[left=2pt,right=2pt]{
    \tau_u = \tau_{\text{RBD}}(q, \dot q, \ddot q, \trbd) + h(q,\dot q, \tnn).
    }
\end{equation}

\subsection{Learn the Lagrangian?}
As seen in Figure~\ref{fig:torque_vectors}, by subtracting from the ``inertial force'' $M\ddot q$ the forces $C(q,\dot q)+G(q)$, we obtain the joint torques $\tau$. While $\tau$ can be obtained via partial differentiation of the Lagrangian as in \eqref{eq:euler-lagrange-2} and observing $\ddot q$, predicting motor torques $\tau_u$ requires the direct identification of $\tau_D$.
If $y:=\tau_u$ such that the target functions becomes
\begin{equation}
    \tau_u = \tau - \tau_{\mathrm{D}},
\end{equation}
then a Lagrangian network would model $\tau_{\mathrm{D}}$ in terms of partial derivatives of the Lagrangian. While it is possible to write a few analytical friction models in terms of potential functions \citep{layton2012principles, xiao2024generalized}
, current works on learning ID as well as our own experimental results show that Lagrangian networks struggle in approximating $\tau_{\mathrm{D}}$ if $\tau_u$ is being measured.

\vspace{2mm}
\subsection{Add model complexity only where required}
To learn dissipative forces alongside Lagrangian networks one can add an additional neural network to the model.
However, increasing the model's representational power by adding a network reduces its data efficiency \citep{von2011statistical}. Therefore, it seems sensible to only add a network to a physics model where it is erroneous. 
If errors in ID solely arise from a miss-specification of $\trbd$, then we can use a fully analytical model such as the DiffNEA and learn its parameters. Similarly, if errors solely arise in $\tau_{\mathrm{D}}$ with $y:=\tau_u$, then approximating these terms via an MLP is viable. Lagrangian networks are useful if some of the unknown functions in the dynamics can be written in terms of a Lagrangian while deriving the Lagrangian analytically is not practical.

\vspace{2mm}
\section{Comparison on an industrial robot}
In this section, we compare learning ID with the different physics-informed networks on data from an industrial robot. A few works also compared Lagrangian networks to other ID models. In particular, \cite{lutter2019lagrangianNNfuruta} compared using DeLaN with an analytical Stribeck friction model to linear system identification showing similar performance. Moreover, \cite{sutanto2020encoding} showed that a DiffNEA with a viscous friction-damping model significantly outperforms DeLaN.
Unlike these prior works, we explicitly look at Lagrangian networks with and without an additional network to learn friction and emphasize that such an additional friction network is indeed required if we do not directly measure the robot's joint torques.

\vspace{2mm}
\subsection{Data collection}
The serial robot used in this work is the MABI MAX 100 of MABI Robotic AG (Figure \ref{fig:overview} left). It consists of six links, weighs approximately 1250 kg, and has a maximum payload of 100 kg. The individual links are each driven by a permanent magnet synchronous motor and contain secondary encoders measuring link positions. The robot is controlled by a Sinumerik 840D sl CNC developed by Siemens.

\subsubsection{Data generation}
To obtain data that is informative about the dynamics, we collected data using a sinusoidal excitation signal that leverages the full joint-specific range of rotation, velocity, and acceleration. 
Based on the approaches of \cite{JIN201521} and \cite{Guo2018DynamicPI}, a third order Fourier series was used for trajectory generation \citep{Lang2016-af}, writing
\begin{equation}
    f(t) = \frac{1}{2} a_0 + \sum_{k=1}^3 (a_k\sin (k \omega t) +b_k\cos (k \omega t)),
\end{equation}
where the parameters $a_0$, $a_k$, and $b_k$ have been chosen manually while the angular frequencies $\omega = \frac{2 \pi}{T}$ with signal period $T$ have been picked such that the joint's actuation starts and ends simultaneously.
To let the robot track the reference trajectories, they were approximated by fifth-degree polynomials. %
The parameters of the approximation are used to program trajectories in the form of geometric code (G-Code) %
\citep{Pott2019}. While the robot follows the reference trajectories, the angles, velocities, and motor currents of each joint are measured with an input-process-output (IPO) cycle of 8 ms.

\subsubsection{Data pre-processing}
As we cannot directly observe joint torques $\tau$, estimates for the motor torques $\tau_u$ were obtained from motor current measurements via \eqref{eq:motor_current_est}.
These control torque estimates were then filtered by a non-causal low-pass filter with the cut-off frequencies $\Vec{f}_{pass} [Hz] =[1.2, 0.3, 0.3, 0.7, 0.5, 0.7]^{T}$ for axes one to six to remove noise perturbations. These values were identified in extensive preliminary tests.
Acceleration estimates are obtained by numerical differentiation of the velocity observations.
The final dataset comprises an input vector consisting of joint angles, velocities, and accelerations, and the output vector being the estimated motor torques.
Prior to training, the data vectors are normalized for each joint dimension to lie in the range $[-1,1]$ and are divided into a 70\% training and  30\% test set, respectively.

\begin{figure}[t]
    \centering
    \includegraphics[width=1\columnwidth]{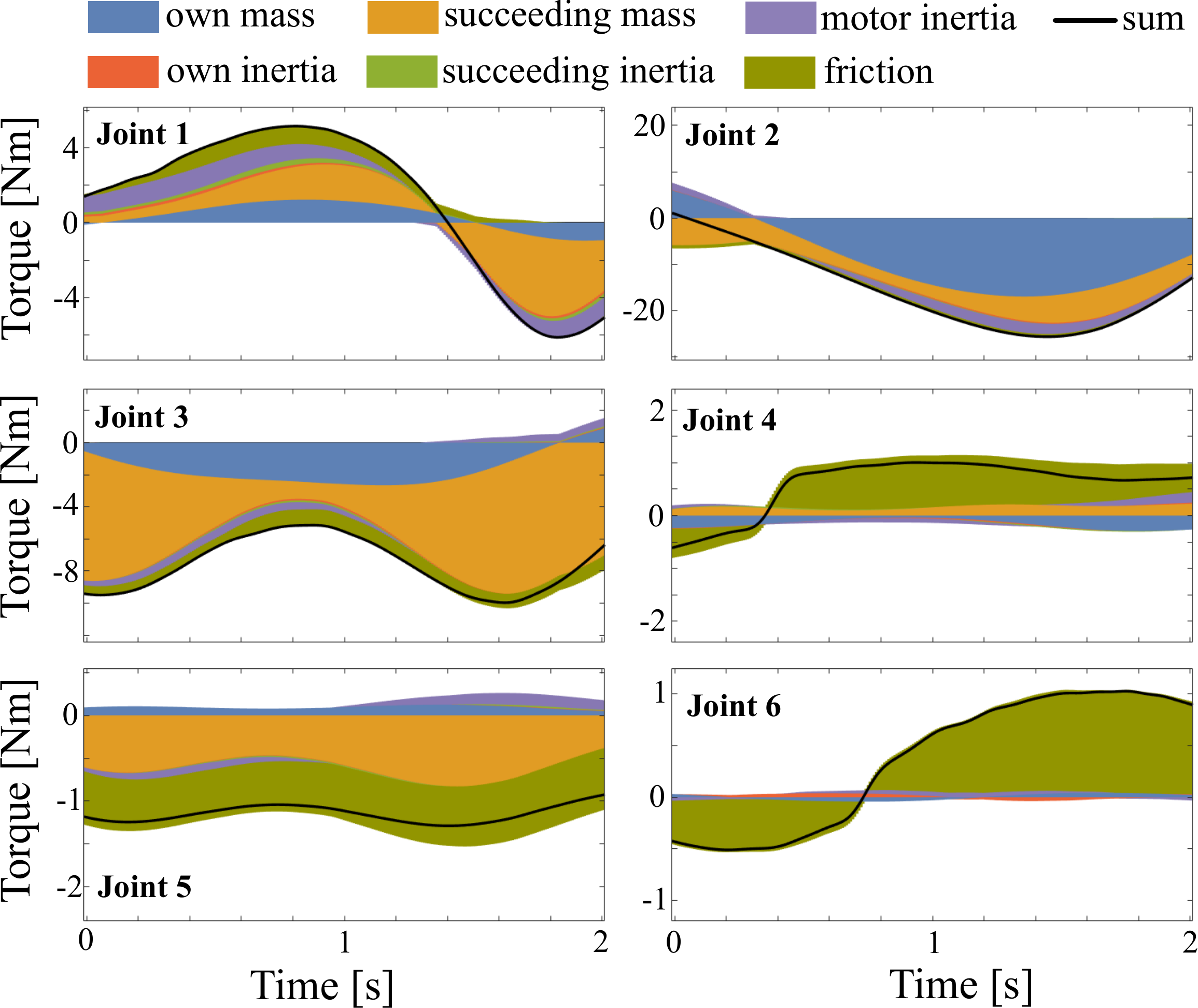}
    \caption{
    The motor torque $\tau_u$ is obtained as the sum of various analytical functions underlying the robot's dynamics.
    }
    \label{fig:share}
\end{figure}

\subsection{Deriving a model baseline via the RNEA}
To set up the Newtonian networks \eqref{eq:newtonian-networks}, we first derive the joint torques \eqref{eq:RBD_model} via the RNEA using the kinematics and inertia parameters provided by the manufacturer. The resulting joint torque vector $\tau_{\text{RBD}}$ does not contain torques due to motor inertia. For the sake of simplicity, we assume that each torque due to the $i$-th motor inertia only affects the $i$-th joint, such that the motor torque reads
\begin{equation} \label{eq:RNEA_mab}
    (\tau_{u})_i = \frac{(\tau_{\text{RBD}})_i}{\psi_{m,i}} + \mathcal{I}_{\text{M},i} \psi_{m,i} \ddot{q}_i + \tau_{\mathrm{D},i}(\dot q_i)
\end{equation}
where $(\tau_{\text{RBD}})_i$ denotes the $i$-th joint torque as obtained from the RNEA, the motor's inertia in rotation direction $\mathcal{I}_{\text{M},i}$, the gear ratio $\psi_{m,i}$, and an analytical viscous friction model $\tau_{\mathrm{D},i}(\dot q_i)= - \theta_{\mathrm{D},i} \dot q_i$. 
While in reality, the torque arising from a motor's inertia also affects the preceding bodies in the kinematic chain, \cite{khalil:cel-02130022} pointed out that the error in the motor torque remains relatively small in the case of high gear ratios. 
As \eqref{eq:RNEA_mab} is linear with respect to the $\theta_{\mathrm{D},i}$, we estimated these parameters via linear regression using the least squares method and the training data set. The resulting model is denoted \emph{RNEA + LQ}.
For the sake of simplicity, we assume that the friction function $\tau_{\mathrm{D},i}$ solely depends on $\dot q_i$ whereas in reality friction also changes with the normal force acting onto the joint axis.

\begin{figure}
    \centering
    \includegraphics[width=1.\columnwidth]{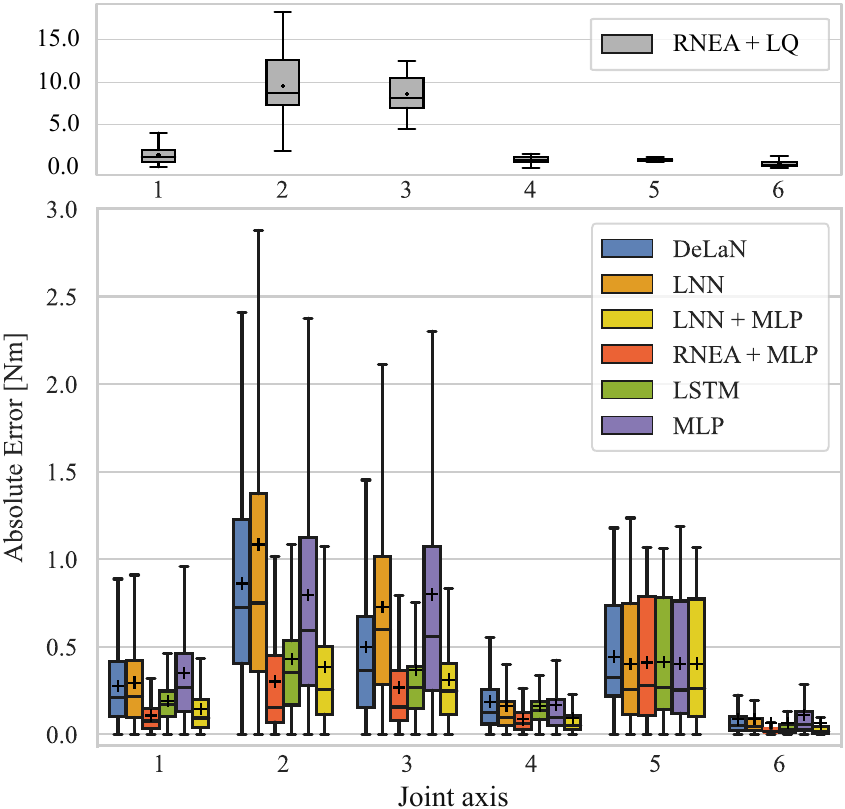}
    \caption{Distribution of absolute test errors for the different models shown for each joint respectively. Even though we do not estimate $\trbd$ of the RNEA, the RNEA/LNN with added MLPs notably outperform the other models.}
    \label{fig:boxplot}
\end{figure}

The physics model \eqref{eq:RNEA_mab} is a sum of different analytical functions. 
Therefore, in Figure  \ref{fig:share}, we show how much the parameter-dependent physics functions contributes to the overall motor torque signal $\tau_u$ (black line). 
The distinction between inertial torques caused by mass and inertia tensors is not directly apparent. After all, the inertia tensor depends on the mass when being described relative to the joint frame instead of the body's center of gravity (Steiner's theorem). In Figure \ref{fig:share}, we refer to a function in \eqref{eq:RNEA_mab} that contains inertia parameters as \emph{inertia} to show the potential influence of the inertia tensors on the computed torque. Function terms that only depend on masses are labeled as \emph{mass}. 
As can be seen in Figure \ref{fig:share}, functions arising from the succeeding mass and friction notably contribute to the required motor torques. 
We note that the friction function in Figure \ref{fig:share} has been calculated as $y_i - (\tau_{\text{RBD}})_i/\psi_{m,i} - \mathcal{I}_{\text{M},i} \psi_{m,i} \ddot{q}_i$ which denotes the difference between estimated motor torques and the model \eqref{eq:RNEA_mab} without friction. As shown in Figure \ref{fig:boxplot}, the friction model $\tau_{\mathrm{D},i}$ differs notably from this estimate which motivates the addition of an MLP to the RNEA.

\begin{table}[t]
\caption{Test RMSE of the data-driven ID models for each robot joint axis. As $y:=\tau_u$, the RMSE of the Lagrangian networks is considerably higher for most joint dimensions compared to the standard neural network and the RNEA+MLP model.}
\centering
\footnotesize
\renewcommand{\arraystretch}{1.1}
\setlength{\tabcolsep}{0.3pt}
\begin{tabular}{@{}ccccccc@{}}
\toprule
\multicolumn{1}{l}{} & \multicolumn{6}{c}{\textbf{Joint axis}} \\
\textbf{Model} & \textbf{1} & \textbf{2} & \textbf{3} & \textbf{4} & \textbf{5} & \textbf{6} \\ \midrule
RNEA+LQ  & \grc{0.3996}{0.0322}{0.40} & \grc{0.5472}{0.0239}{0.55} & \grc{0.7402}{0.033}{0.741} & \grc{0.6132}{0.077}{0.614} & \grc{0.7011}{0.4287}{0.7011} & \grc{0.5887}{0.1374}{0.5887} \\
DeLaN    & \grc{0.0773}{0.0322}{0.13} & \grc{0.0561}{0.0239}{0.12} & \grc{0.0565}{0.033}{0.126} & \grc{0.1723}{0.077}{0.1742} & \grc{0.4388}{0.4287}{0.4475} & \grc{0.1469}{0.1374}{0.1918} \\
LNN      & \grc{0.1291}{0.0322}{0.13} & \grc{0.1115}{0.0239}{0.12} & \grc{0.1258}{0.033}{0.126} & \grc{0.1529}{0.077}{0.1742} & \grc{0.4287}{0.4287}{0.4475} & \grc{0.1436}{0.1374}{0.1918} \\
LNN+MLP  & \grc{0.0436}{0.0322}{0.40} & \grc{0.0338}{0.0239}{0.55} & \grc{0.0384}{0.033}{0.741} & \grc{0.1093}{0.077}{0.614} & \grc{0.4357}{0.4287}{0.7011} & \grc{0.1487}{0.1374}{0.5887} \\
RNEA+MLP & \grc{0.0322}{0.0322}{0.13} & \grc{0.0239}{0.0239}{0.12} & \grc{0.0331}{0.033}{0.126} & \grc{0.0773}{0.077}{0.1742} & \grc{0.4379}{0.4287}{0.4475} & \grc{0.1654}{0.1374}{0.1918} \\
LSTM     & \grc{0.0509}{0.0322}{0.13} & \grc{0.0405}{0.0239}{0.12} & \grc{0.0605}{0.033}{0.126} & \grc{0.1364}{0.077}{0.1742} & \grc{0.4348}{0.4287}{0.4475} & \grc{0.1375}{0.1374}{0.1918} \\
MLP      & \grc{0.1045}{0.0322}{0.13} & \grc{0.0563}{0.0239}{0.12} & \grc{0.0945}{0.033}{0.126} & \grc{0.1741}{0.077}{0.1742} & \grc{0.4383}{0.4287}{0.4475} & \grc{0.1918}{0.1374}{0.1918} \\ \bottomrule
\end{tabular}
\normalsize
\label{tab:Multi_case_trajectory_params}
\vspace{2mm}
\end{table}

\subsection{Setup of data-driven models}
As a comparison baseline, we resort to an MLP and a long short-term memory (LSTM) network. The LSTM has 2 layers of width 128 and was trained on a sequence length of 20. The MLP has 2 layers of width 64 with Tanh activations. MLPs have been the standard comparison baseline in works on Lagrangian neural networks \citep{lutter2018deep, cranmer2020lagrangian}. For the Lagrangian networks, we resort to the code bases provided alongside the respective papers of DeLaN \citep{lutter2018deep} and LNN \citep{cranmer2020lagrangian}. We add an additional MLP to the LNN which we refer to as \emph{LNN + MLP}. We compare the aforementioned models to models based on the Newton-Euler equations of motion, namely the analytical RNEA model \eqref{eq:RNEA_mab} and a Newtonian network which we obtained by replacing $\tau_{\mathrm{D},i}$ in \eqref{eq:RNEA_mab} by an MLP. The latter model we refer to as \emph{RNEA + MLP}. In the RNEA models, $\trbd$ remains fixed during training. 

\subsection{Comparison of prediction accuracy}
In what follows, we train different ID models on the data of the robot for which $y:=\tau_u$.
Before evaluation, an extensive hyper-parameter search for all models was carried out in the AI platform \textit{Weights \& Biases} using GPU \citep{wandb}. All models were trained using Adam \citep{kingma2014adam} and did not show any convergence or stability issues.

Figure \ref{fig:boxplot} shows the models' absolute errors on the test data for each joint respectively. The Newtonian network (\emph{RNEA+MLP}) significantly outperforms all other models and in particular the Lagrangian networks (\emph{DeLaN} and \emph{LNN}). Especially the robot's first four joints are subject to large friction forces that arise from the weight of its bodies pressing on its joints as shown in Figure \ref{fig:overview} (left). In comparison, the robot's last two joints (five and six) are subject to small friction forces. We hypothesize that due to these relatively small forces, the difference in performance between Newtonian and Lagrangian networks is less pronounced at joints five and six. Table \ref{tab:Multi_case_trajectory_params} shows the model's root mean square error (RMSE) for each joint respectively. Also in terms of the RMSE, the RNEA+MLP and LSTM outperform the Lagrangian networks.

\begin{figure}[t]
\vspace{1.5mm}
    \centering
    \includegraphics[width=0.8\columnwidth]{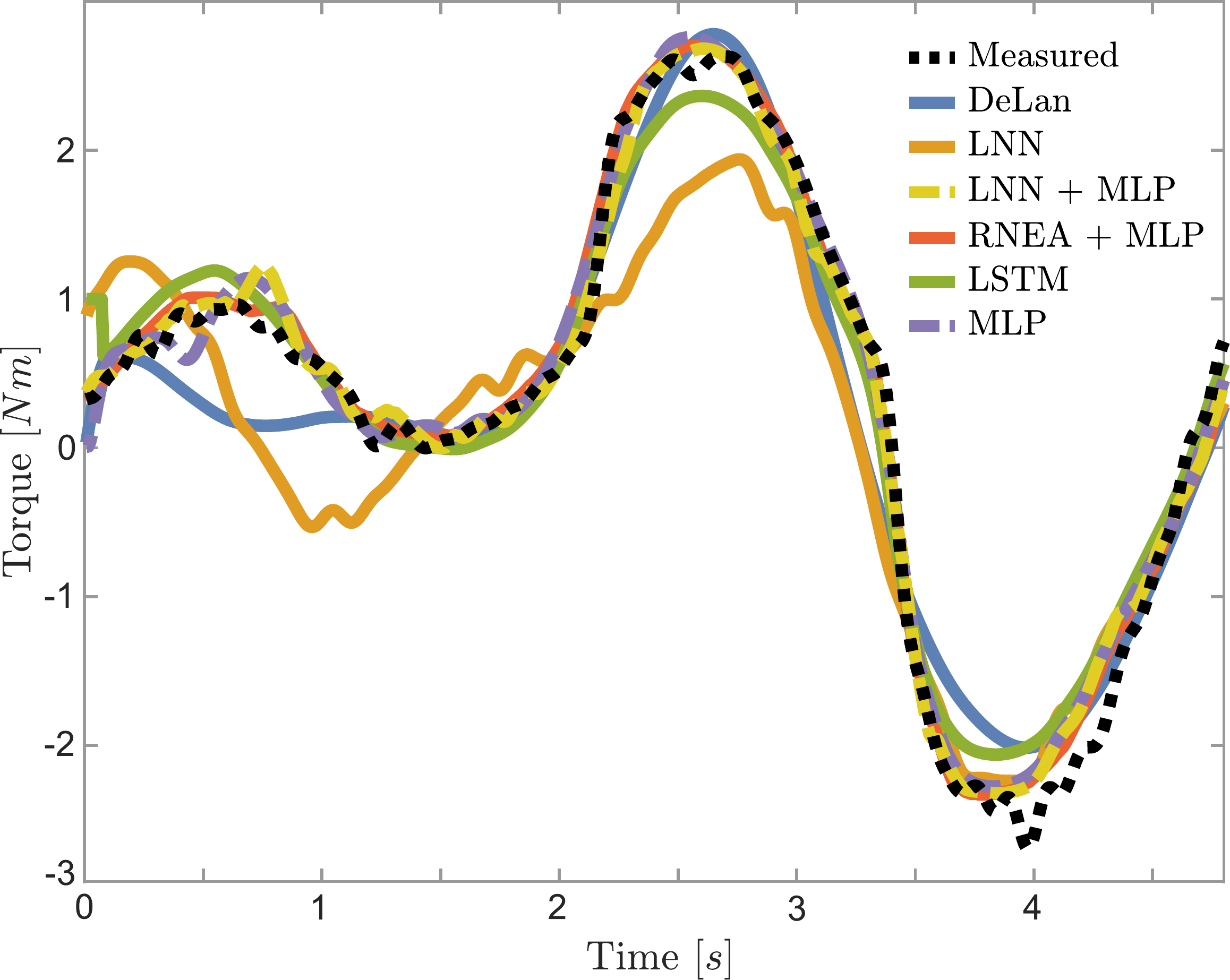}
    \vspace{1.5mm}
    \caption{Dissipative torque estimates over a test data trajectory of axis 1 for the different inverse dynamics models. These estimates where obtained by subtracting the networks prediction of $\tau_u$ from an RNEA model baseline predicting $\tau$.  %
    }
    \label{fig:comparison}
\end{figure}

Figure \ref{fig:comparison} shows the prediction of the trained ID models on a test data trajectory which is representative for the general behavior. While the ``vanilla'' Lagrangian networks yield good approximations for some parts of the functions, their performance is notably improved through the addition of an MLP which are again outperformed by the RNEA + MLP.

\vspace{2mm}
\section{Conclusion}
We investigated the effectiveness of Newtonian and Lagrangian networks in learning the inverse dynamics of an industrial robot. While the DeLaN \citep{lutter2019lagrangianNNfuruta} and the LNN \citep{cranmer2020lagrangian} are promising ID models when joint torques are being directly measured, when motor torques are measured instead, Lagrangian models require the addition of an MLP to perform well. Moreover in the case of measurements $y$ being solely the motor torques $\tau_u$, we showed that an analytically derived RNEA model with an added MLP may notably outperform Lagrangian networks. 
An interesting question for future research is how to best initialize the weights of Lagrangian networks. While in the case of Newtonian networks it is straightforward to estimate MLP parameters via gradient-based optimization, the identification of physics parameters alongside neural network parameters remains challenging. Finally, for real-time application in robotic systems, an evaluation of the computational efficiencies of the models using different types of robots crucial.

\section*{Acknowledgment}
Funded by the Deutsche Forschungsgemeinschaft (DFG, German Research Foundation) under Germany’s Excellence Strategy – EXC-2023 Internet of Production – 390621612.

\bibliography{literature}

\end{document}

%% file: packages.tex
\usepackage[x11names, xcdraw, table]{xcolor}
\usepackage{times}
\usepackage{amsfonts}
\usepackage{comment}
\usepackage{caption}
\usepackage{amsmath}
\usepackage[most]{tcolorbox} 
\usepackage{booktabs}
\usepackage{graphicx}

\definecolor{ourblue}{rgb}{0.368,0.507,0.71}
\definecolor{ourgreen}{rgb}{0.56,0.692,0.195}
\definecolor{ourviolet}{rgb}{0.528,0.471,0.701}

\usepackage{natbib}

\makeatletter
\let\old@ssect\@ssect %
\makeatother

\usepackage{hyperref}
\hypersetup{
	colorlinks=true,       %
	linkcolor=ourblue,        %
	citecolor=ourblue,        %
	filecolor=ourblue,     %
	urlcolor=ourviolet
}

\makeatletter
\def\@ssect#1#2#3#4#5#6{%
  \NR@gettitle{#6}%
  \old@ssect{#1}{#2}{#3}{#4}{#5}{#6}%
}
\makeatother

\usepackage{csquotes}

%% file: commands.tex
\newcommand{\transp}{\mathrm{\mathsf{T}}}  %

\newcommand{\eg}{\emph{e.g.},{ }}

\newcommand*{\Rn}{\mathbb{R}^n}

\newcommand*{\trbd}{\theta_{\text{RBD}}}

\newcommand*{\tnn}{\theta_{\text{NN}}}

\newcommand*{\lnn}{LNN}

\tcbset{boxrule=0.4pt, colback=black!0!white, colframe=black, 
        highlight math style= {enhanced, %
            colframe=black,colback=black!0!white,boxsep=-5pt}
        }

\definecolor{high}{HTML}{FFFFFF}%
\definecolor{low}{HTML}{8FB032}%
\definecolor{top}{HTML}{FFFFFF}%
\newcommand*{\opacity}{90}%

\newcommand{\grc}[3]{
    \ifdimcomp{#1pt}{>}{#3 pt}{#1}{
        \ifdimcomp{#1pt}{<}{#2 pt}{#1}{
            \pgfmathparse{int(round(100*((#1-#2)/(#3-#2))))}
            \xdef\tempa{\pgfmathresult}
            \cellcolor{high!\tempa!low!\opacity} #1
    }}
}